\documentclass[conference]{IEEEtran}

\IEEEoverridecommandlockouts                              

\usepackage{amsmath} 
\usepackage{amssymb}  
\usepackage{amsmath,amssymb,amsfonts}
\usepackage{bm} 
\usepackage[T1]{fontenc}
\usepackage{graphicx}
\usepackage{flafter} 
\usepackage{placeins} 
\usepackage{booktabs}
\usepackage{multirow}
\usepackage{pifont}

\usepackage[T1]{fontenc}
\usepackage{comment} 

\newif\ifanonymous
\anonymousfalse       

\title{\vspace*{0.25in}\Large \bf
Structure-Aware Robust Fine-Tuning: Defending Vision-Language-Action Robots\\
Against Physical Attention Hijacking
}

\ifanonymous
\author{Anonymous Author(s)\\
Anonymous Institution\\
{\ttfamily anonymous@anonymous.com}}
\else
\author{
Jinquan Zhang$^{1,2}$,
Dongfu Yin$^{1,*}$,
Run Yang$^{1,2}$,
Yufeng Yan$^{1,2}$,
Zhen Tian$^{1}$,
F. Richard Yu$^{3}$%
\thanks{This work was supported in part by the Research Task Assignment Project from Guangdong Laboratory of Artificial Intelligence and Digital Economy (SZ) under Grant No. GML-26420004, and in part by the Shenzhen Science and Technology Program under Grant No. KJZD20240903104400001.}%

\thanks{$^{1}$Guangdong Laboratory of Artificial Intelligence and Digital Economy (SZ), Shenzhen, China.}%
\thanks{$^{2}$Shenzhen University, Shenzhen, China.}%
\thanks{$^{3}$Carleton University, Ottawa, Canada.}%
\thanks{$^{*}$Corresponding author: Dongfu Yin, yindongfu@gml.ac.cn.}%
}
\fi

\begin{document}

\bstctlcite{BSTcontrol}

\maketitle

\thispagestyle{empty}
\pagestyle{empty}

\begin{abstract}
Vision-Language-Action (VLA) policies promise general robotic manipulation, but their robustness against physical-world attacks remains fragile. In particular, we show that physically realizable adversarial patches can reliably induce failures by triggering a mechanism we call policy-critical action-to-vision attention hijacking, where action-conditioned attention is diverted from task-relevant regions to a localized patch. To demonstrate the threat, we propose Attention-Guided Semantic Disruption (AGSD), an Expectation-over-Transformation (EOT) optimized printable patch that jointly (i) concentrates action-to-vision attention on the patch and (ii) disrupts vision--language semantic alignment, yielding strong cross-task and cross-architecture transfer. To mitigate such attacks, we introduce Structure-Aware Robust Fine-Tuning (SARF), a zero-inference-overhead defense that fine-tunes only the visual encoder using feature anchoring, policy-critical attention correction, and language-guided geometric consistency restricted to semantically relevant regions. On LIBERO, SARF reduces OpenVLA's failure rate under AGSD from 100\% to 14.2--56.8\% (28.6\% avg.) across suites while preserving clean performance, and on a real PiPER manipulator it improves average success under AGSD from 23.0\% to 65.0\%. These results highlight mechanism-level robustness as a practical path to securing VLA robots against physical attention hijacking.
\end{abstract}

\section{INTRODUCTION}

\begin{figure}[!t]
  \centering
  \includegraphics[width=\columnwidth]{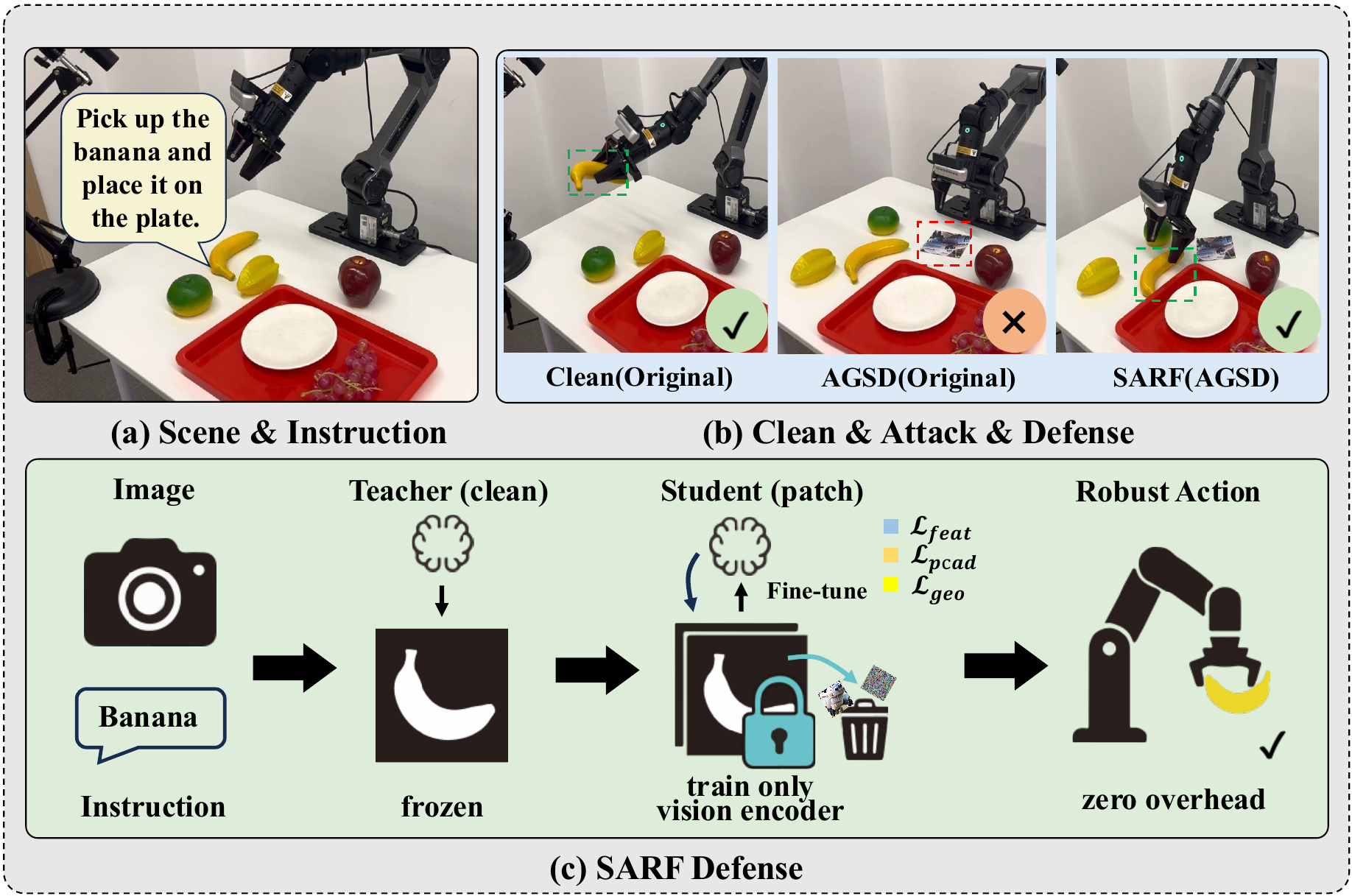}
\caption{Motivation and overview of SARF against physical attention hijacking.
(a) PiPER tabletop manipulation conditioned on a natural-language instruction.
(b) The original VLA succeeds on clean inputs but fails under a printable AGSD patch; SARF restores success under the same attack.
(c) SARF fine-tunes only the visual encoder in a teacher--student framework using feature anchoring ($\mathcal{L}_{\text{feat}}$), policy-critical action-token attention distillation ($\mathcal{L}_{\text{pcad}}$), and language-guided geometric consistency ($\mathcal{L}_{\text{geo}}$), with zero inference overhead.}
  \label{fig:fig1}
\end{figure}

Vision--Language--Action (VLA) policies map images and natural-language instructions directly to low-level actions and have enabled increasingly general robotic manipulation \cite{rt2,openvla,octo,pi0}. 
Because VLAs close the perception--action loop end-to-end, localized visual failures can immediately translate into erroneous physical behaviors, making robustness a first-order safety concern.

A practical and low-cost threat is the physically realizable adversarial patch: a printable localized pattern that remains effective under viewpoint and illumination changes when optimized with Expectation-over-Transformation (EOT) \cite{advpatch,eot}. 
Recent studies report that patch attacks can substantially degrade VLA manipulation in both simulation and real settings \cite{mft,evavla,vla_vuln}. 
Existing defenses (e.g., the defense accompanying EDPA) largely rely on adversarial fine-tuning to align \emph{global} representations between clean and patched observations \cite{edpa_vla}. 
However, manipulation often depends on \emph{sparse} task-critical evidence (e.g., end-effector and target object), suggesting that robustness may hinge on stabilizing \emph{where} the policy attends, not only \emph{what} it represents.

We identify a failure mode that is particularly damaging for VLA control: \emph{policy-critical action-to-vision attention hijacking}. 
In modern VLAs, action outputs are driven by cross-attention from a small set of action-query tokens to visual tokens. 
A patch can become a strong attention attractor, capturing attention mass from these action queries and suppressing attention on task-relevant regions, thereby derailing long-horizon trajectories and motivating defenses that stabilize action-conditioned attention pathways (Fig.~\ref{fig:fig1}).

Guided by this observation, we propose (i) a stress-test attack to expose the mechanism and (ii) a targeted defense. On the attack side, we introduce Attention-Guided Semantic Disruption (AGSD), an EOT-optimized printable patch that jointly concentrates action-to-vision attention on the patch and disrupts vision–language semantic alignment, improving transfer across tasks and architectures. To mitigate this vulnerability, we propose Structure-Aware Robust Fine-Tuning (SARF), a zero-inference-overhead teacher--student robust fine-tuning framework that updates only the visual encoder while freezing the multimodal backbone and action head.
SARF combines feature anchoring, policy-critical action-token attention distillation to counteract hijacking, and language-guided geometric consistency restricted to semantically relevant regions.

Our contributions are threefold:
\begin{itemize}
  \item  We present AGSD, an EOT-optimized printable patch that targets policy-critical action-to-vision attention, jointly inducing attention hijacking and semantic misalignment to achieve strong cross-task and cross-architecture transfer.
  \item We propose SARF, which achieves robust fine-tuning with zero inference overhead by combining feature anchoring, policy-critical attention correction, and language-guided geometric consistency restricted to semantically relevant regions.
  \item We evaluate our methods on LIBERO \cite{libero} and a PiPER tabletop setup, demonstrating the broad vulnerability of multiple VLA architectures (e.g., $\pi_0$ and OpenVLA variants) to AGSD, and showing that SARF consistently enhances the robustness of OpenVLA across diverse physical patch scenarios while preserving clean performance.
\end{itemize}

\FloatBarrier 

\begin{figure*}[!t]
  \centering
  \includegraphics[width=\textwidth]{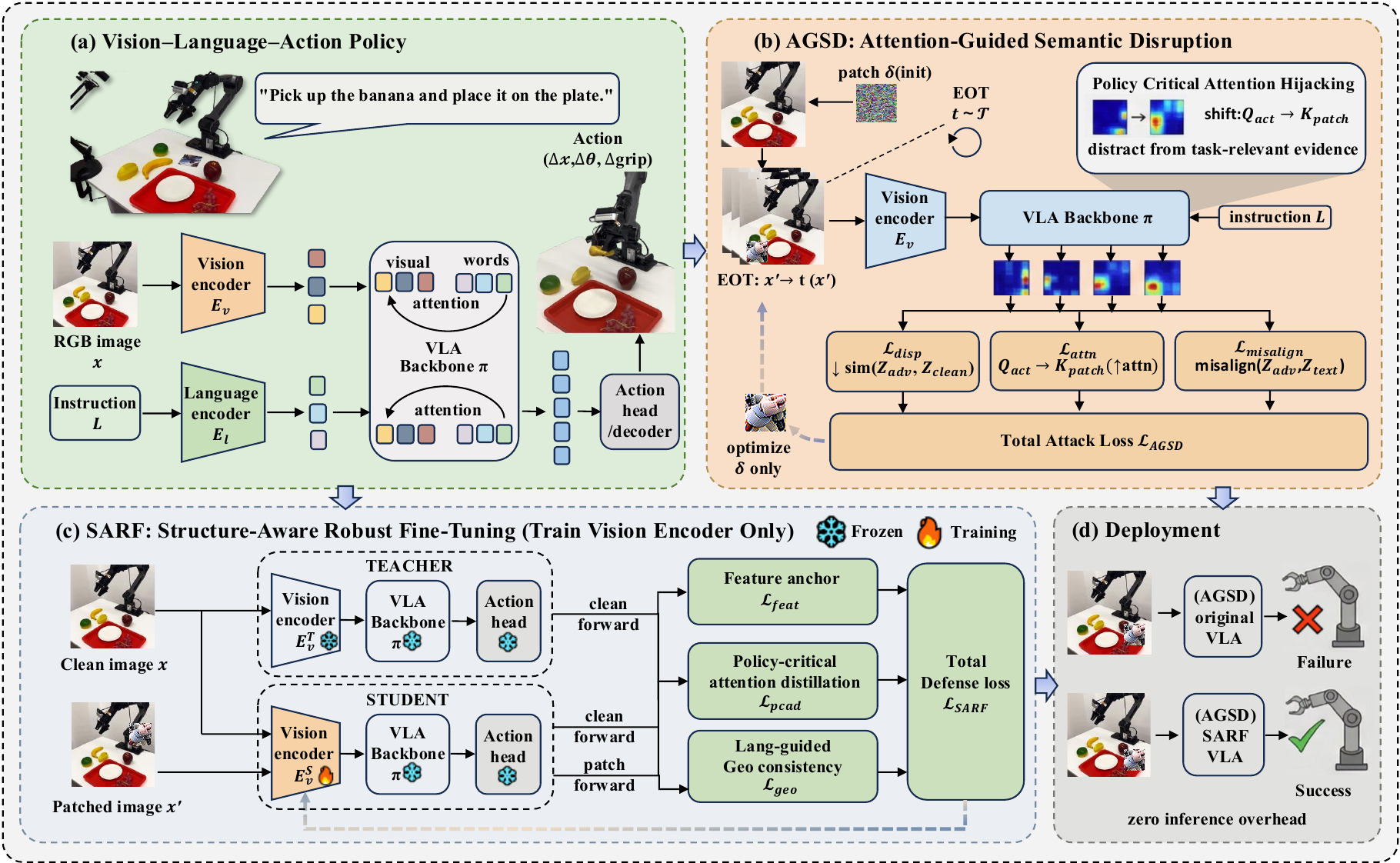}
  \caption{Physical attention hijacking and our attack--defense pipeline.
(a) A VLA policy encodes an RGB observation $\mathbf{x}$ and instruction $L$ and produces low-level actions via cross-attention.
(b) AGSD optimizes a printable patch $\boldsymbol{\delta}$ under EOT to hijack \emph{action-to-vision} attention and disrupt vision--language alignment.
(c) SARF performs teacher--student robust fine-tuning that updates only the student visual encoder $E_v^S$ (freezing $\pi$ and the action head) using feature anchoring $\mathcal{L}_{feat}$, policy-critical attention distillation $\mathcal{L}_{pcad}$, and language-guided geometric consistency $\mathcal{L}_{geo}$ .
(d) At deployment, SARF restores successful execution under the same patch with zero inference overhead.}
  \label{fig:fig2}
\end{figure*}

\section{Related Work}

\subsection{VLA / Robot Foundation Policies for Manipulation}
Vision--Language--Action (VLA) policies enable robots to execute open-vocabulary manipulation by conditioning low-level control on images and natural-language instructions.
RT-2 demonstrates that scaling vision--language knowledge can benefit robotic control \cite{rt2}, while OpenVLA provides an open-source foundation VLA trained on large-scale robot demonstrations \cite{openvla}.
Octo further advances open generalist robot policies across diverse robots and sensors \cite{octo}, and $\pi_0$ explores flow-based action generation for general robot control \cite{pi0}.
Complementary lines investigate alternative action modeling and cross-modal fusion for robotic perception \cite{diffpolicy,cai2025mft_fusion,cai2025dstr}.
However, current foundation policies primarily prioritize scaling and zero-shot generalization, leaving their vulnerabilities to physical adversarial threats largely underexplored. 

\subsection{Physical Patch Attacks and Robustness Evaluation for VLA Systems}
Printable adversarial patches, optimized via Expectation-over-Transformation (EOT) \cite{advpatch,eot}, pose a realistic physical threat whose perception errors are often amplified by closed-loop robotic execution.
Recent evaluation efforts highlight that VLA robustness under physical variations and adversarial conditions remains underexplored and can degrade dramatically under patch attacks \cite{mft,evavla}.
Beyond general robustness evaluation, VLA-specific adversarial studies propose patch objectives tailored to robotic action spaces, including UADA/UPA-style attacks that directly induce action-level discrepancies \cite{vla_vuln}.
EDPA constructs model-agnostic patches by increasing representation discrepancy and image--text misalignment, and proposes a corresponding defense via adversarial fine-tuning \cite{edpa_vla}.
More recent works emphasize transferability and universality across models and sim-to-real conditions (e.g., UPA-RFAS) \cite{upa_rfas}, propose attention-guided sparse attacks for efficiency and stealthiness (e.g., ADVLA) \cite{advla}, and provide broader benchmarking suites for adversarial and backdoor threats across the VLA lifecycle (e.g., AttackVLA) \cite{attackvla}.
Despite these advances, existing physical attacks mostly target global feature disruption or text-to-vision misalignment, overlooking the specific attention pathways that drive VLA action generation. To expose this blind spot, our AGSD explicitly exploits policy-critical action-to-vision attention combined with semantic disruption, achieving strong transferability.

\subsection{Defenses and Robust Fine-Tuning Against Patch Attacks}
Classical defenses against adversarial perturbations include adversarial training \cite{madry,trades}, data augmentation and domain randomization \cite{domainrand}, and input-time purification using generative models \cite{diffpure,defensegan,pixeldefend}.
For localized patch threats, PatchGuard provides provable robustness by combining small receptive fields with robust masking \cite{patchguard}.
However, purification-style defenses often introduce inference overhead and must be carefully evaluated against adaptive attacks to avoid a false sense of security \cite{obfg}.
In the VLA setting, EDPA's defense improves robustness by adversarially fine-tuning the visual encoder to better align clean and patched representations \cite{edpa_vla}, but it does not explicitly stabilize the \emph{attention mechanism} that determines where the policy looks when producing actions.
Meanwhile, robust/continual adaptation methods aim to preserve pretrained capabilities during fine-tuning, e.g., via regularization or distillation \cite{ewc,lwf}, attention transfer \cite{attntransfer}, or parameter merging to improve robustness without extensive retraining \cite{retain_merge}.
Crucially, prior defenses in VLA settings either introduce inference overhead or focus solely on aligning global representations, failing to stabilize the fundamental attention mechanisms that determine action outputs. To overcome this, our SARF framework achieves zero-inference-overhead robustness by directly correcting policy-critical attention and enforcing localized geometric consistency.

\section{Methodology}
\label{sec:method}

We formalize the physical threat model, introduce AGSD, and then present SARF for improving VLA robustness against physical attention hijacking.

\subsection{Attention-Guided Semantic Disruption (AGSD)}
\label{subsec:threat}

We model the physical attack as a printable, localized adversarial patch optimization problem.
Given the original RGB observation $\mathbf{x}\in\mathbb{R}^{H\times W\times 3}$,
the patched adversarial input $\mathbf{x}'$ is defined as
\begin{equation}
\mathbf{x}' = (\mathbf{1} - \mathbf{m}) \odot \mathbf{x} + \mathbf{m} \odot \boldsymbol{\delta},
\label{eq:patch}
\end{equation}
where $\boldsymbol{\delta}$ denotes the adversarial patch,
$\mathbf{m}\in\{0,1\}^{H\times W}$ is a binary spatial mask indicating the patch location,
and $\odot$ is the element-wise product.

The attacker aims to find an optimal patch $\boldsymbol{\delta}^*$ that remains effective under common physical imaging variations.
We adopt the Expectation over Transformation (EOT) framework:
\begin{equation}
\boldsymbol{\delta}^* =
\operatorname*{argmin}_{\boldsymbol{\delta}}\;
\mathbb{E}_{t \sim \mathcal{T}}
\left[
\mathcal{L}_{\text{AGSD}}\big(t(\mathbf{x}')\big)
\right],
\label{eq:eot}
\end{equation}
where $t(\cdot)$ is sampled from a transformation distribution $\mathcal{T}$
(e.g., random rotation, translation, and perspective tilting),
and the expectation encourages robustness in real-world deployment.

To effectively attack a VLA model, we define the attack total loss as a weighted combination of three terms:
\begin{equation}
\mathcal{L}_{\text{AGSD}} =
\lambda_{\text{attn}} \mathcal{L}_{\text{attn}}
- \lambda_{\text{disp}} \mathcal{L}_{\text{disp}}
- \lambda_{\text{misalign}} \mathcal{L}_{\text{misalign}},
\label{eq:AGSD}
\end{equation}
where $\lambda_{\text{attn}}$, $\lambda_{\text{disp}}$, and $\lambda_{\text{misalign}}$ are balancing coefficients.
Note that $\mathcal{L}_{\text{disp}}$ and $\mathcal{L}_{\text{misalign}}$ are subtracted in
Eq.~\eqref{eq:AGSD}, meaning that minimizing $\mathcal{L}_{\text{AGSD}}$ effectively \emph{maximizes} these two losses,
thus disrupting feature consistency and image--text alignment.
We define each component as follows.

\subsubsection{ Attention Guidance Loss ($\mathcal{L}_{\text{attn}}$)}
This loss directly manipulates the cross-modal attention mechanism to ``hijack'' the model's perceptual focus.
It forces the policy-critical action queries to over-attend to the patch region, thereby diverting action-to-vision cross-attention away from task-relevant evidence:
\begin{equation}
\mathcal{L}_{\text{attn}} =
-\frac{1}{|\mathcal{Q}_{\text{act}}||K_{\text{patch}}|}
\sum_{q\in \mathcal{Q}_{\text{act}}}\sum_{k\in K_{\text{patch}}}
\bar{A}_{q,k},
\label{eq:lattn}
\end{equation}
where $\mathcal{Q}_{\text{act}}$ denotes the set of policy action-query tokens that condition the action head,
$K_{\text{patch}}$ is the set of visual key tokens inside the patch region (determined by $\mathbf{m}$),
and $\bar{A}_{q,k}$ is the mean cross-attention weight aggregated from the last three cross-attention layers.
The leading negative sign maximizes the attention weight from the action queries to the adversarial patch.

\subsubsection{ Feature Dispersion Loss ($\mathcal{L}_{\text{disp}}$)}
To destabilize semantic features, we adopt an InfoNCE-style objective:
\begin{equation}
\mathcal{L}_{\text{disp}} =
-\log
\frac{\exp(\text{sim}(\mathbf{z}_{\text{adv}}, \mathbf{z}_{\text{clean}}) / \tau_{\text{nce}})}
{\sum_{j=1}^{B} \exp(\text{sim}(\mathbf{z}_{\text{adv}}, \mathbf{z}_{\text{clean}}^{(j)}) / \tau_{\text{nce}})},
\label{eq:ldisp}
\end{equation}
where $\mathbf{z}_{\text{adv}}$ and $\mathbf{z}_{\text{clean}}$ denote the embeddings of the adversarial and clean images,
$B$ is the batch size, $\tau_{\text{nce}}$ is the temperature, and $\text{sim}(\cdot,\cdot)$ is cosine similarity.
Since $\mathcal{L}_{\text{disp}}$ is subtracted in Eq.~\eqref{eq:AGSD}, the optimizer effectively maximizes it,
which reduces the similarity between $\mathbf{z}_{\text{adv}}$ and $\mathbf{z}_{\text{clean}}$ and induces strong feature-space perturbations.

\subsubsection{ Image--Text Misalignment Loss ($\mathcal{L}_{\text{misalign}}$)}
This loss aims to maximize image--text misalignment in the joint embedding space:
\begin{equation}
\mathcal{L}_{\text{misalign}} =
\frac{1}{B} \sum_{i=1}^{B}
\left\|
\text{sim}(\mathbf{z}_{\text{adv}}^{(i)}, \mathbf{z}_{\text{text}}^{(i)})
-
\text{sim}(\mathbf{z}_{\text{clean}}^{(i)}, \mathbf{z}_{\text{text}}^{(i)})
\right\|_1,
\label{eq:lmis}
\end{equation}
where $\mathbf{z}_{\text{text}}^{(i)}$ is the embedding of the $i$-th text instruction.
By maximizing this $L_1$ distance (implicitly via Eq.~\eqref{eq:AGSD}), the attacker forces the image--text matching of the adversarial image
to deviate significantly from the correct alignment of the clean image.

\subsection{Proposed Defense: Structure-Aware Robust Fine-Tuning (SARF)}
\label{subsec:defense}

To defend against the above patch attack, we propose SARF.
SARF fine-tunes only the visual encoder of the VLA model and uses a Teacher model pretrained on clean data to guide the recovery of a perturbed Student model.
The overall SARF objective consists of feature anchoring, attention rectification, and geometric consistency:
\begin{equation}
\mathcal{L}_{\text{SARF}} =
\lambda_{\text{feat}} \mathcal{L}_{\text{feat}}
+ \lambda_{\text{pcad}} \mathcal{L}_{\text{pcad}}
+ \lambda_{\text{geo}} \mathcal{L}_{\text{geo}}.
\label{eq:lsarf}
\end{equation}

\subsubsection{ Feature Anchor Loss ($\mathcal{L}_{\text{feat}}$)}
To prevent catastrophic forgetting during fine-tuning, we enforce the Student's high-level semantic features
to remain directionally consistent with the Teacher:
\begin{equation}
\mathcal{L}_{\text{feat}} =
1 - \frac{1}{N} \sum_{i=1}^{N}
\frac{\mathbf{z}_{S}^{(i)} \cdot \mathbf{z}_{T}^{(i)}}
{\|\mathbf{z}_{S}^{(i)}\| \|\mathbf{z}_{T}^{(i)}\|},
\label{eq:lfeat}
\end{equation}
where $N$ is the number of visual patch tokens, and $\mathbf{z}_{S}^{(i)}$ and $\mathbf{z}_{T}^{(i)}$
are the $i$-th patch-level feature vectors from the Student and Teacher, respectively.
This loss anchors the feature space by minimizing the cosine distance $(1-\cos\theta)$.

\subsubsection{ Policy-Critical Attention Distillation ($\mathcal{L}_{\text{pcad}}$)}
Motivated by the VLA policy structure, we distill attention only for Action Tokens that determine action outputs.
We align Student and Teacher attention distributions using a symmetric Jensen--Shannon Divergence (JSD) objective:
%
\begin{equation}
\label{eq:lpcad}
\mathcal{L}_{\text{pcad}}
=
\frac{1}{H\lvert\mathcal{Q}_{\text{act}}\rvert}
\sum_{h=1}^{H}\sum_{q\in\mathcal{Q}_{\text{act}}}
D_{\text{JS}}\!\left(P_T^{(h)} \parallel P_S^{(h)}\right),
\end{equation}
where $H$ is the number of heads and $\mathcal{Q}_{\text{act}}$ indexes Action tokens.
For each $h$ and $q\in\mathcal{Q}_{\text{act}}$, we form attention distributions over visual tokens
$P^{(h)}(k\mid q)=\mathrm{Softmax}(A^{(h)}_{q,k}/\tau_{\text{attn}})$ and minimize
$D_{\text{JS}}(P_T^{(h)}(\cdot\mid q)\parallel P_S^{(h)}(\cdot\mid q))$ to align policy-critical action-to-vision attention.

\subsubsection{ Language-Guided Geometric Consistency ($\mathcal{L}_{\text{geo}}$)}
To preserve the geometry of task-relevant objects while suppressing background noise, we introduce a language-guided geometric mask.
We first compute the importance of each patch $i$ using Teacher text-to-vision attention:
$m_i = \max_{q \in \mathcal{Q}_{\text{txt}}} A_{q,i}^{T}$,
and construct a sharpened pairwise mask
$\mathbf{M}_{ij}=(m_i \cdot m_j)^2$.
Since this mask is derived from the clean Teacher stream rather than the perturbed Student stream, it provides a stable reference for task-relevant regions.

Based on this mask, the geometric loss is defined as
\begin{equation}
\mathcal{L}_{\text{geo}} =
\frac{\sum_{i,j} \mathbf{M}_{ij} \cdot \left(G_{ij}^{S} - G_{ij}^{T}\right)^2}
{\sum_{i,j} \mathbf{M}_{ij} + \epsilon},
\label{eq:lgeo}
\end{equation}
where the Gram element $G_{ij}$ captures pairwise feature relations (texture and structure):
\begin{equation}
G_{ij} = \frac{\mathbf{z}^{(i)} \cdot \mathbf{z}^{(j)}}{\|\mathbf{z}^{(i)}\| \|\mathbf{z}^{(j)}\|}.
\label{eq:gram}
\end{equation}
This loss constrains only the regions activated by $\mathbf{M}$, enabling structure denoising on task-critical objects
(e.g., the end-effector and manipulated objects) while automatically down-weighting background perturbations.

\section{Experiments}
\label{sec:exp}

We first demonstrate that physical patch attacks are feasible and can severely degrade Vision--Language--Action (VLA) robots.
We then analyze the underlying failure mechanism via attention-level diagnostics,
propose SARF as a targeted solution, and validate robustness both in simulation and on a real PiPER system.
Unless otherwise specified, results are reported on four LIBERO suites (\textit{Spatial, Object, Goal, Long}).

\subsection{Experimental Setup}
\label{subsec:exp_setup}

\subsubsection{Benchmarks}
We evaluate on four LIBERO simulation task suites (reporting Failure Rate, FR) and a real-world PiPER tabletop setup (reporting Success Rate, SR). For PiPER, we conduct 100 independent trials per condition with randomized initial object poses, camera viewpoints, and distances to assess physical robustness.

\subsubsection{Models}
We evaluate representative VLA policies, including OpenVLA and its fine-tuned variant OpenVLA-oft,
as well as a stronger foundation-policy baseline $\pi_0$ (Pi0).
All models are evaluated using their default action interfaces and inference settings to ensure fair comparison.

\subsubsection{Attacks}
We consider localized, printable adversarial patches and compare:
(i) Clean (no patch),
(ii) Random (non-optimized printable patch),
(iii) UADA and UPA from~\cite{vla_vuln},
(iv) EDPA from~\cite{edpa_vla},
and (v) AGSD (ours).
Unless otherwise stated, all methods use the same patch size and placement distribution.

\subsubsection{Defenses}
We report three settings on OpenVLA:
Original (no defense),
AF (Defense Method in the EDPA),
and SARF (ours).
SARF fine-tunes only the \emph{visual encoder} and distills \emph{policy-critical} action-to-vision attention from a clean Teacher model,
with language-guided geometric consistency restricted to semantically relevant regions. We instantiate SARF on OpenVLA because its action-conditioned visual attention is accessible; extension to diffusion- or flow-style decoders requires identifying analogous policy-critical visual pathways.

\subsubsection{Metrics}
For LIBERO, we report Failure Rate (FR, \%).
For the real robot, we report Success Rate (SR, \%).
When evaluating attack effectiveness (e.g., AGSD), a higher FR indicates a stronger attack (worse robustness).
When evaluating defense effectiveness (e.g., SARF), a lower FR indicates better robustness while preserving performance on clean inputs.

\subsubsection{Implementation details and EOT settings}
To ensure stable optimization by balancing inherent gradient magnitudes and prioritizing critical objectives, we set the AGSD coefficients to $\lambda_{\text{attn}}=0.8$, $\lambda_{\text{disp}}=0.2$, and $\lambda_{\text{misalign}}=0.5$. These values are selected such that the attention-guidance loss acts as the primary driver of optimization, while the remaining terms provide auxiliary semantic disruption. For the SARF defense, we set $\lambda_{\text{feat}}=0.5$, $\lambda_{\text{pcad}}=1.0$, and $\lambda_{\text{geo}}=0.3$. This configuration reflects our core design principle: prioritizing the distillation of policy-critical attention ($\mathcal{L}_{\text{pcad}}$) to counteract hijacking, while leveraging $\mathcal{L}_{\text{feat}}$ and $\mathcal{L}_{\text{geo}}$ as structure-aware constraints to preserve the pretrained visual backbone's integrity. In LIBERO, the adversarial patch occupies 5\% of the image area. On the real PiPER robot, we evaluate printed physical patches with sizes of 20$\times$20 cm, 15$\times$15 cm, 8$\times$8 cm, and 5$\times$5 cm. To ensure physical realizability, all attacks are optimized and evaluated under an Expectation-over-Transformation (EOT) framework that incorporates random in-plane rotation ($\theta \sim \mathcal{U}(-30^\circ,30^\circ)$), translation (up to 10\% of the image dimensions), scaling ($s \sim \mathcal{U}(0.9,1.1)$), random patch placement (uniformly sampled within the image bounds), and mild perspective/illumination jitter. SARF uses the same data budget as AF and updates only the visual encoder; the deployed architecture and inference latency remain unchanged.

\FloatBarrier

\begin{figure}[!htbp]
  \centering
  \includegraphics[width=\columnwidth]{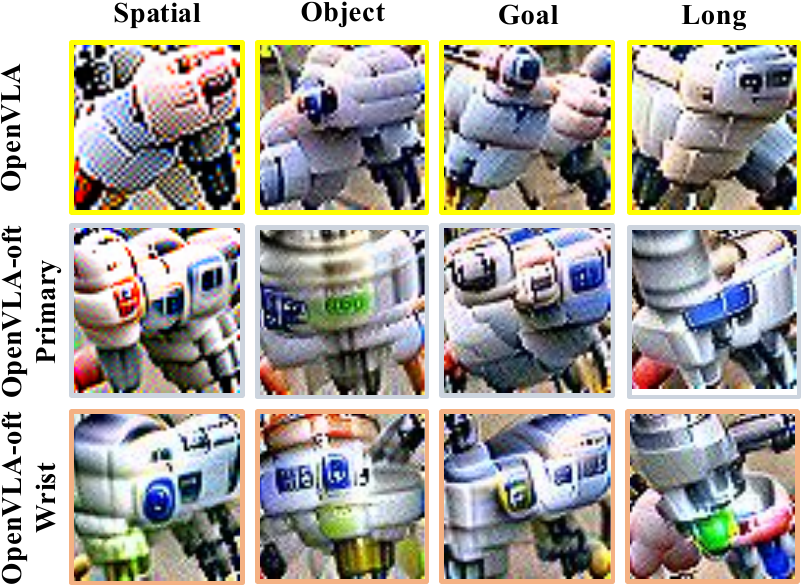}
  \caption{Printable AGSD patch examples.
Representative AGSD patches optimized under the same EOT pipeline.
Columns correspond to the four LIBERO suites (\textit{Spatial}, \textit{Object}, \textit{Goal}, \textit{Long}); rows show patches optimized for OpenVLA and OpenVLA-oft using primary and wrist camera observations.
These patches are used in our physical patch-attack evaluations.}
  \label{fig:fig3_Patch_Visualization}
\end{figure}

\subsection{Physical Attention Hijacking}
\label{subsec:threat_story}

\subsubsection{Quantitative results of attack effectiveness}
As shown in Table~\ref{tab:attack_fr_models}, even non-optimized \emph{Random} patches already increase failure rates compared to \emph{Clean} inputs, indicating that localized physical perturbations can impair manipulation.
However, optimized patches are substantially more damaging; \emph{AGSD} drives \emph{OpenVLA} to 100\% failure across all suites, improving over \emph{Random} by 24.4--64.2 points. Crucially, AGSD transfers beyond the attacked model: on \emph{OpenVLA-oft}, failure rises from 10.0--32.0\% (Random) to 93.6--100\% (AGSD), and on the stronger $\pi_0$ baseline from 5.2--48.6\% (Random) to 48.8--80.2\% (AGSD).
Relative to EDPA, AGSD is comparable on OpenVLA while remaining markedly stronger on OpenVLA-oft and $\pi_0$, supporting cross-architecture transfer.
Consistently, Table~\ref{tab:openvla_attack_defense} (Original column) confirms that, under the same evaluation protocol used for defense studies, the undefended OpenVLA exhibits near-certain failure under AGSD across all suites.

\begin{table}[!htbp]
  \centering
  \caption{Attack effectiveness on LIBERO (Failure Rate, \%).
Failure rates (mean$\pm$std) of OpenVLA, OpenVLA-oft, and $\bm{\pi_0}$ under Clean, Random, and optimized patch attacks (EDPA, AGSD) across four suites.
Higher FR indicates stronger attacks; best (highest) results among optimized attacks are in \textbf{bold}.}
  \label{tab:attack_fr_models}
  \footnotesize
  \setlength{\tabcolsep}{5pt}
  \renewcommand{\arraystretch}{1.15}
  \begin{tabular}{l l c c c}
    \toprule
    \multirow{2}{*}{\textbf{Suite}} & \multirow{2}{*}{\textbf{Method}} & \multicolumn{3}{c}{\textbf{Failure Rate (FR$\uparrow$)}} \\
    \cmidrule(lr){3-5}
    & & \textbf{OpenVLA} & \textbf{OpenVLA-oft} & $\bm{\pi_0}$ \\
    \midrule
    \multirow{3}{*}{\textbf{Spatial}}
      & Clean        & 14.2 $\pm$ 0.5  & 2.4 $\pm$ 0.4  & 3.4 $\pm$ 0.3 \\
      & Random       & 35.8 $\pm$ 1.3  & 10.0 $\pm$ 1.1 & 6.0 $\pm$ 0.8 \\
      & EDPA\cite{edpa_vla}         & \textbf{100 $\pm$ 0.0}   & 39.7 $\pm$ 0.9 & 29.8 $\pm$ 1.6 \\
      & AGSD (Ours)  & \textbf{100 $\pm$ 0.0}   & \textbf{97.2 $\pm$ 0.6} & \textbf{48.8 $\pm$ 2.1} \\
    \midrule
    \multirow{3}{*}{\textbf{Object}}
      & Clean        & 11.6 $\pm$ 0.4  & 2.6 $\pm$ 0.3  & 2.0 $\pm$ 0.2 \\
      & Random       & 44.6 $\pm$ 1.2  & 20.4 $\pm$ 1.5 & 5.2 $\pm$ 0.5 \\
      & EDPA\cite{edpa_vla}         & \textbf{100 $\pm$ 0.0}   & 52.3 $\pm$ 0.8 & 39.5 $\pm$ 1.7 \\
      & AGSD (Ours)  & \textbf{100 $\pm$ 0.0}   & \textbf{93.6 $\pm$ 1.2} & \textbf{50.4 $\pm$ 2.4} \\
    \midrule
    \multirow{3}{*}{\textbf{Goal}}
      & Clean        & 20.8 $\pm$ 1.5  & 3.0 $\pm$ 0.6  & 10.4 $\pm$ 1.1 \\
      & Random       & 42.0 $\pm$ 1.2  & 16.2 $\pm$ 1.3 & 16.8 $\pm$ 1.4 \\
      & EDPA\cite{edpa_vla}         & \textbf{100 $\pm$ 0.0}   & 80.8 $\pm$ 0.4 & 44.3 $\pm$ 2.0 \\
      & AGSD (Ours)  & \textbf{100 $\pm$ 0.0}   & \textbf{100 $\pm$ 0.0}  & \textbf{70.8 $\pm$ 1.8} \\
    \midrule
    \multirow{3}{*}{\textbf{Long}}
      & Clean        & 46.2 $\pm$ 2.0  & 4.8 $\pm$ 0.7   & 42.0 $\pm$ 1.6 \\
      & Random       & 75.6 $\pm$ 2.4  & 32.0 $\pm$ 2.1  & 48.6 $\pm$ 1.9 \\
      & EDPA\cite{edpa_vla}         & \textbf{100  $\pm$ 0.0}   & 86.4 $\pm$ 1.9  & 70.7 $\pm$ 1.6 \\
      & AGSD (Ours)  & \textbf{100  $\pm$ 0.0}   & \textbf{100 $\pm$ 0.0}   & \textbf{80.2 $\pm$ 1.5} \\
    \bottomrule
  \end{tabular}
\end{table}

\begin{figure}[!htbp]
  \centering
  \includegraphics[width=\linewidth]{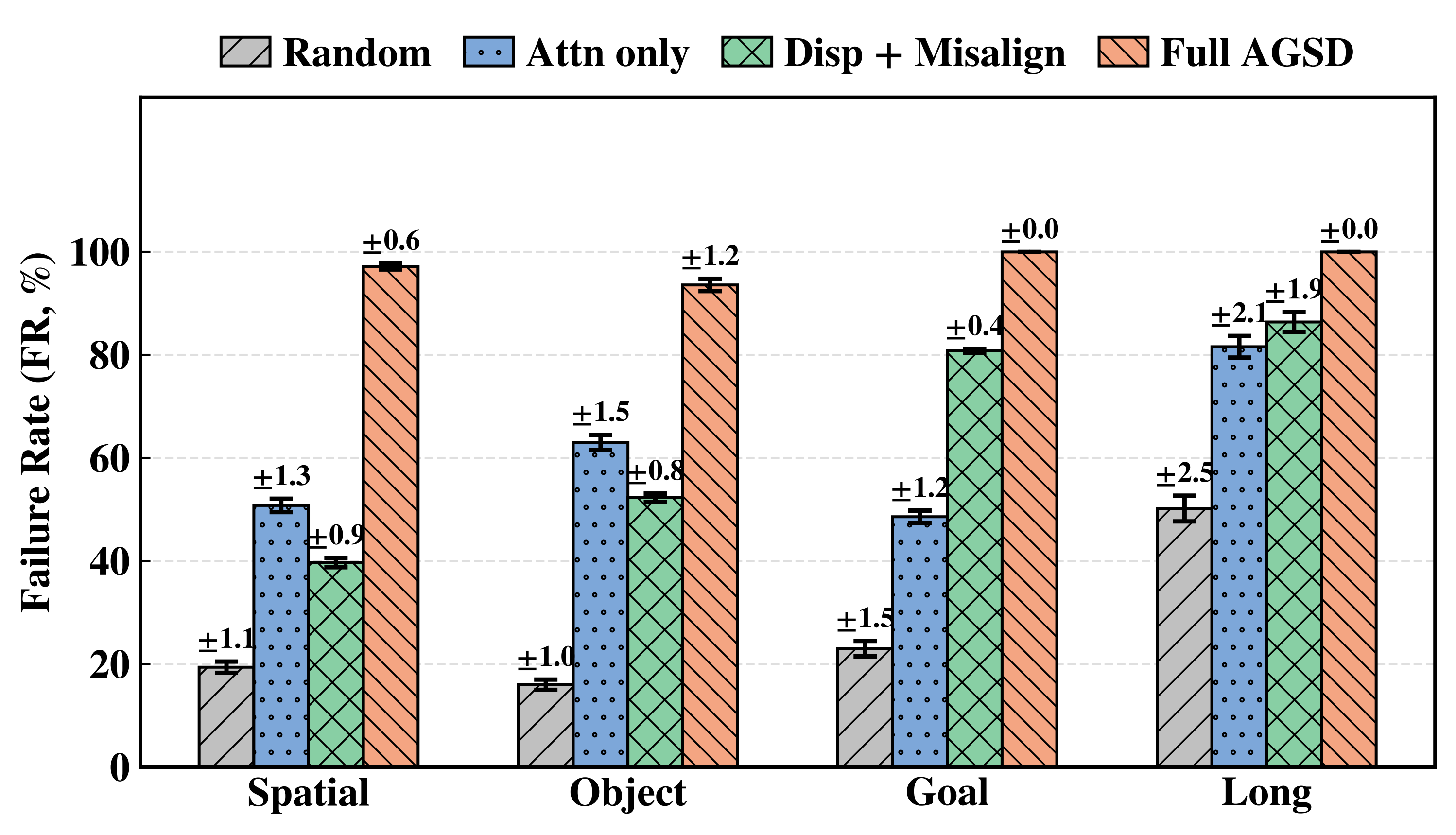}
  \caption{AGSD objective ablation.
Failure rates (FR, \%) of OpenVLA on four LIBERO suites under four printable patch variants: Random, Attn-only, Disp+Misalign, and Full AGSD.
Higher FR indicates a stronger attack; error bars denote variability across repeated evaluations.}
  \label{fig:fig4_agsd_ablation}
\end{figure}

\subsubsection{AGSD produces printable and deployable patches}
To illustrate physical plausibility, Fig.~\ref{fig:fig3_Patch_Visualization} visualizes representative \emph{printable} adversarial patches optimized by AGSD against OpenVLA and OpenVLA-oft under the same EOT transformation pipeline.
Across suites and viewpoints (primary and wrist cameras), the optimized patterns are concrete and reproducible,
supporting real-world deployment rather than relying on imperceptible digital noise.

\subsubsection{Mechanistic insight via objective ablation}
To understand \emph{why} AGSD is particularly effective, we conduct an objective-level ablation in Fig.~\ref{fig:fig4_agsd_ablation}.
Optimizing only the attention-guidance term (Attn-only) already increases FR, indicating that explicitly manipulating
\emph{policy-critical action-to-vision attention} can drive failures.
Optimizing semantic disruption without explicit attention guidance (Disp + Misalign) also degrades performance by destabilizing visual features and vision--language alignment.
Crucially, the Full AGSD objective that couples attention guidance with semantic disruption achieves the strongest attacks across all LIBERO suites,
suggesting that these components are complementary and together yield more transferable physical patches.

\subsubsection{Qualitative attention hijacking under AGSD}
While Table~\ref{tab:attack_fr_models} and Fig.~\ref{fig:fig4_agsd_ablation} establish the strength and contributing factors of AGSD,
we further provide qualitative attention evidence in Fig.~\ref{fig:fig5_sarf_mech}.
Under AGSD, the Original policy's action-to-vision attention repeatedly collapses onto the patch region across rollout timesteps,
diverting focus away from task-relevant evidence.
This motivates defenses that explicitly stabilize policy-critical attention pathways.

\FloatBarrier

\subsection{SARF for Stabilizing Policy-Critical Attention}
\label{subsec:defense_story}

\subsubsection{Goal and evaluation protocol}
We evaluate SARF as a \emph{drop-in} defense against physically realizable patch attacks:
it introduces \emph{zero inference overhead} (no test-time modules and unchanged policy interface) and updates only the visual encoder during training.

\begin{table}[!htbp]
  \centering
  \caption{OpenVLA robustness under physical patch attacks (FAILURE
RATE,\%).
Failure rates (mean$\pm$std) on four LIBERO suites for Original, AF(Defense Method in the EDPA), and SARF under Clean, Random, and optimized attacks (UADA/UPA/EDPA/AGSD); lower is better and best is in bold.
SARF--AGSD uses an adaptive (re-optimized) AGSD patch against the SARF-tuned model.}
  \label{tab:openvla_attack_defense}
  \scriptsize
  \setlength{\tabcolsep}{5pt}
  \renewcommand{\arraystretch}{1.10}
  \begin{tabular}{l l c c c}
    \toprule
    \textbf{Suite} & \textbf{Method} &
    \multicolumn{3}{c}{\textbf{Failure Rate (FR$\downarrow$)}} \\
    \cmidrule(lr){3-5}
    & & \textbf{Original} & \textbf{AF\cite{edpa_vla}} & \textbf{SARF (Ours)} \\
    \midrule
    \multirow{6}{*}{\textbf{Spatial}}
      & Clean        &\textbf{14.2 $\pm$ 0.5} & 17.9 $\pm$ 0.8 & 14.4 $\pm$ 0.6 \\
      & Random       & 35.8 $\pm$ 1.3 & 19.4 $\pm$ 1.1 & \textbf{15.0 $\pm$ 0.7} \\
      & UADA\cite{vla_vuln}         & 98.8 $\pm$ 0.2 & 65.4 $\pm$ 2.5 & \textbf{16.2 $\pm$ 0.9} \\
      & UPA\cite{vla_vuln}          & 99.0 $\pm$ 0.1 & 46.6 $\pm$ 2.1 & \textbf{15.8 $\pm$ 0.8} \\
      & EDPA\cite{edpa_vla}         & 100 $\pm$ 0.0  & 39.4 $\pm$ 1.8 & \textbf{16.5 $\pm$ 1.0} \\
      & \textbf{AGSD (Ours)}  & 100 $\pm$ 0.0  & 90.2 $\pm$ 1.5 & \textbf{17.0 $\pm$ 1.1} \\
    \midrule
    \multirow{6}{*}{\textbf{Object}}
      & Clean        & \textbf{11.6 $\pm$ 0.4} & 17.3 $\pm$ 0.9 & 11.8 $\pm$ 0.5 \\
      & Random       & 44.6 $\pm$ 1.2 & 16.0 $\pm$ 1.0 & \textbf{12.4 $\pm$ 0.6} \\
      & UADA\cite{vla_vuln}         & 92.0 $\pm$ 1.5 & 58.8 $\pm$ 2.3 & \textbf{13.2 $\pm$ 0.8} \\
      & UPA\cite{vla_vuln}          & 94.2 $\pm$ 1.1 & 43.9 $\pm$ 2.0 & \textbf{12.8 $\pm$ 0.7} \\
      & EDPA\cite{edpa_vla}         & 100 $\pm$ 0.0  & 58.6 $\pm$ 2.4 & \textbf{13.6 $\pm$ 0.9} \\
      & \textbf{AGSD (Ours)}  & 100 $\pm$ 0.0  & 99.8 $\pm$ 0.2 & \textbf{14.2 $\pm$ 1.0} \\
    \midrule
    \multirow{6}{*}{\textbf{Goal}}
      & Clean        & \textbf{20.8 $\pm$ 1.5} & 22.8 $\pm$ 1.2 & 21.0 $\pm$ 1.4 \\
      & Random       & 42.0 $\pm$ 1.2 & 23.0 $\pm$ 1.5 & \textbf{22.2 $\pm$ 1.6} \\
      & UADA\cite{vla_vuln}         & 98.6 $\pm$ 0.4 & 91.6 $\pm$ 1.8 & \textbf{24.6 $\pm$ 1.7} \\
      & UPA\cite{vla_vuln}          & 96.4 $\pm$ 0.8 & 68.3 $\pm$ 2.2 & \textbf{23.8 $\pm$ 1.5} \\
      & EDPA\cite{edpa_vla}         & 100 $\pm$ 0.0  & 73.9 $\pm$ 2.5 & \textbf{25.4 $\pm$ 1.8} \\
      & \textbf{AGSD (Ours)}  & 100 $\pm$ 0.0  & 97.2 $\pm$ 0.6 & \textbf{26.5 $\pm$ 2.0} \\
    \midrule
    \multirow{6}{*}{\textbf{Long}}
      & Clean        & \textbf{46.2 $\pm$ 2.0} & 49.0 $\pm$ 2.1 & 46.6 $\pm$ 1.9 \\
      & Random       & 75.6 $\pm$ 2.4 & 50.2 $\pm$ 2.5 & \textbf{48.8 $\pm$ 2.2} \\
      & UADA\cite{vla_vuln}         & 99.8 $\pm$ 0.2 & 97.4 $\pm$ 0.8 & \textbf{53.2 $\pm$ 2.6} \\
      & UPA\cite{vla_vuln}          & 99.8 $\pm$ 0.1 & 86.7 $\pm$ 1.5 & \textbf{51.5 $\pm$ 2.4} \\
      & EDPA\cite{edpa_vla}         & 100 $\pm$ 0.0  & 91.2 $\pm$ 1.2 & \textbf{54.0 $\pm$ 2.7} \\
      & \textbf{AGSD (Ours)}  & 100 $\pm$ 0.0  & 99.0 $\pm$ 0.5 & \textbf{56.8 $\pm$ 2.9} \\
    \bottomrule
  \end{tabular}
\end{table}

\begin{figure}[!htbp]
  \centering
  \includegraphics[width=\columnwidth]{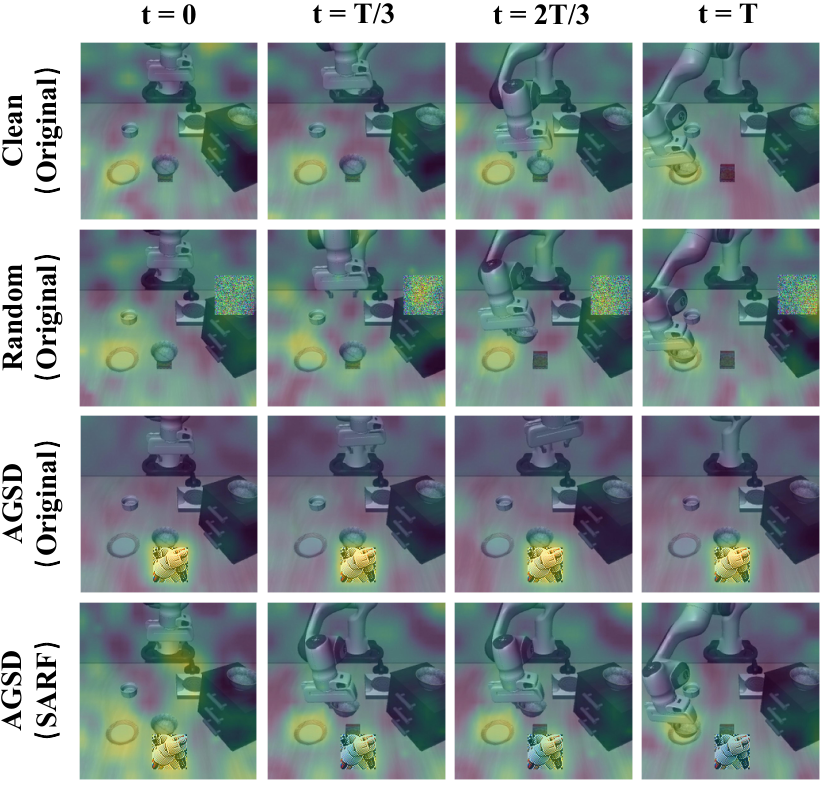}
  \caption{AGSD hijacks policy-critical attention, while SARF restores task-relevant focus.
Action-to-vision cross-attention heatmaps overlaid on the RGB observation at four rollout timesteps ($t\!=\!0,\,T/3,\,2T/3,\,T$).
Rows show Clean/Random baselines (Original), AGSD-attacked Original, and AGSD-attacked SARF.
AGSD causes attention to collapse onto the patch region across time, whereas SARF suppresses patch fixation and re-centers attention on task-relevant evidence.}
  \label{fig:fig5_sarf_mech}
\end{figure}

\subsubsection{Main robustness gains under strong transferable patches}
Table~\ref{tab:openvla_attack_defense} reports OpenVLA failure rates (FR; lower is better) under diverse physical patch attacks.
Under the strongest stress-test attack, AGSD, the \emph{undefended} policy fails almost always (FR$=100\%$) on all four suites.
In contrast, SARF reduces FR to 17.0\% / 14.2\% / 26.5\% / 56.8\% on \textit{Spatial/Object/Goal/Long},
corresponding to absolute drops of 83.0 / 85.8 / 73.5 / 43.2 points, respectively.
Averaged across suites, SARF decreases FR from 100.0\% to 28.6\% under AGSD (a 71.4-point reduction).
The higher failure rate on the \textit{Long} suite suggests that minor residual attention jitter can compound over long-horizon trajectories, though the improvement over the undefended baseline remains substantial.

\subsubsection{Consistent improvements across attack baselines}
Beyond AGSD, SARF yields consistent robustness gains across prior patch baselines.
Averaged over suites, FR decreases from 97.3\% (UADA) to 26.8\%, from 97.4\% (UPA) to 26.0\%, and from 100.0\% (EDPA) to 27.4\%.
These results indicate that SARF improves robustness broadly under diverse optimization objectives and transfer settings.

\subsubsection{Comparison to adversarial fine-tuning (AF)}
Compared with AF (the EDPA defense), SARF achieves markedly lower FR under strong attacks.
For example, under AGSD, AF remains highly vulnerable with an average FR of 96.6\%, whereas SARF reduces it to 28.6\%
(an additional 67.9-point reduction over AF).
Similar margins hold for EDPA (65.8\% to 27.4\%) and UADA (78.3\% to 26.8\%) on average.
This massive performance gap highlights that for end-to-end visuomotor policies, stabilizing \textbf{where} the model looks is far more effective than broadly aligning \textbf{what} the model represents.

\subsubsection{Robustness under an adaptive attacker}
Importantly, the AGSD results for SARF are obtained using an \emph{adaptive} setting where the patch is re-optimized against the frozen SARF-tuned model using the same AGSD objective and EOT settings.
Even under this stronger threat model, SARF maintains substantially lower FR than both the undefended policy and AF.

\subsubsection{Clean performance is preserved}
Robustness gains do not come at the cost of clean-task degradation:
on clean inputs, FR changes only marginally from 14.2/11.6/20.8/46.2\% (Original) to 14.4/11.8/21.0/46.6\% (SARF) across suites
(average 23.2\% to 23.5\%).

\subsubsection{Mechanism-level evidence: attention is re-centered under attack}
Fig.~\ref{fig:fig5_sarf_mech} provides an attention-level diagnosis consistent with the quantitative improvements.
Under AGSD, the original policy exhibits persistent patch fixation where action-to-vision cross-attention collapses onto the patch region across rollout timesteps.
After SARF, attention is re-centered on semantically relevant regions under the \emph{same} physical patch, supporting the mechanism-targeted explanation that stabilizing policy-critical attention helps prevent long-horizon failures.

\FloatBarrier

\subsection{Real-Robot Validation on PiPER}
\label{subsec:real_robot}

\subsubsection{Protocol and metric}
We evaluate three real-world manipulation tasks on a PiPER tabletop platform (Table~\ref{tab:real_robot}) and report Success Rate (SR, \%). For each task, we run 100 physical trials with randomized initial object poses and randomized trial order. We report clean performance for the undefended policy (Original), and evaluate robustness under a printed AGSD patch by comparing Original, AF~\cite{edpa_vla}, and SARF (ours). To reflect realistic sensing variations, we vary the camera viewpoint and distance across trials.

\begin{figure}[!htbp]
  \centering
  \includegraphics[width=\columnwidth]{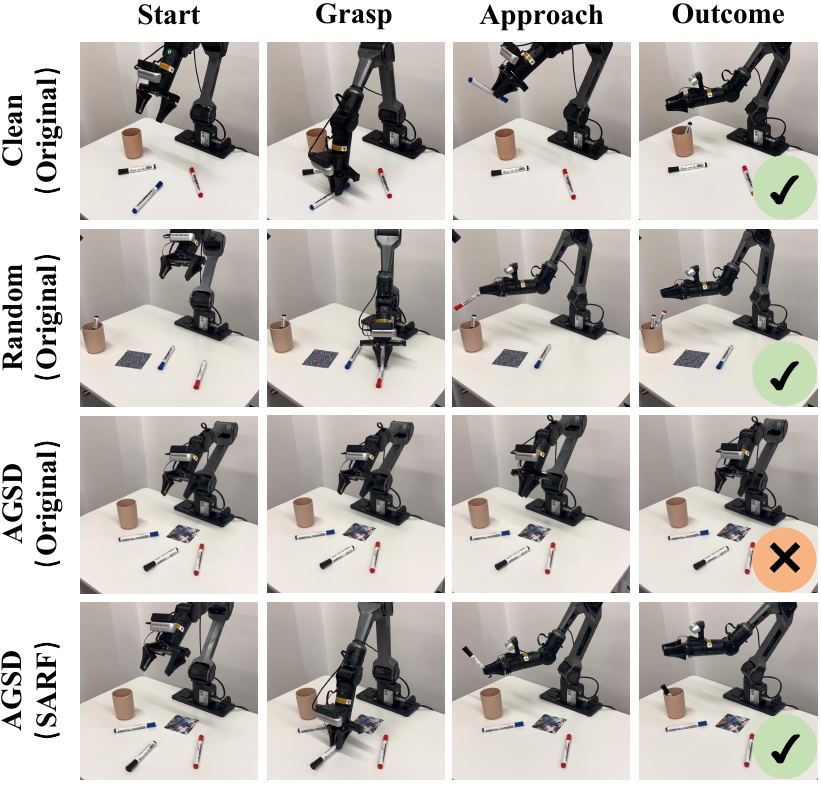}
  \caption{Real-robot qualitative comparison under physical patch attacks (PiPER tabletop).
Four keyframes (Start/Grasp/Approach/Outcome) under Clean and Random baselines (Original), and under an AGSD patch with the Original and SARF policies.
The AGSD patch causes failure for the original policy, while SARF restores successful execution under the same printed patch (zero inference overhead).}
  \label{fig:fig6_real_robot_attn}
\end{figure}

\begin{table}[!htbp]
  \centering
  \caption{Real-robot results on PiPER under a printed AGSD patch (Success Rate, \%). Success rates are averaged over 100 physical trials per condition. We compare the undefended policy (Original), the strongest simulation baseline (AF), and our SARF. SARF decisively restores manipulation capabilities under the same physical patch.}
  \label{tab:real_robot}
  \footnotesize
  \setlength{\tabcolsep}{6pt}
  \renewcommand{\arraystretch}{1.15}
  \begin{tabular}{lcccc}
    \toprule
    \multirow{2}{*}{\textbf{Task}} & \textbf{Clean} & \multicolumn{3}{c}{\textbf{Under AGSD Attack}} \\
    \cmidrule(lr){3-5}
    & \textbf{Original} & \textbf{Original} & \textbf{AF \cite{edpa_vla}} & \textbf{SARF (Ours)} \\
    \midrule
    pick \& place & 79.0 & 32.0 & 51.0 & \textbf{74.0} \\
    open drawer   & 71.0 & 23.0 & 42.0 & \textbf{63.0} \\
    stack         & 66.0 & 14.0 & 29.0 & \textbf{58.0} \\
    \midrule
    \textbf{Average}       & 72.0 & 23.0 & 40.7 & \textbf{65.0} \\
    \bottomrule
  \end{tabular}
\end{table}

\subsubsection{SARF recovers real-world performance under a printed AGSD patch}
As shown in Table~\ref{tab:real_robot}, the printed AGSD patch substantially degrades real-world performance, reducing the average success rate of the undefended policy (Original) from 72.0\% (clean) to 23.0\% under attack. The simulation baseline AF~\cite{edpa_vla} partially improves robustness (40.7\% on average), whereas SARF restores manipulation much more effectively under the \emph{same} printed patch, achieving 65.0\% average success (a 42.0-point gain over the attacked Original). The gains are consistent across tasks: for \textit{pick \& place}, SARF reaches 74.0\% under attack, close to the clean performance of 79.0\%; for the precision-heavy \textit{stack} task, SARF improves success from 14.0\% to 58.0\% (a 44.0-point gain), although an 8.0-point gap to the clean setting remains. These results suggest that SARF mitigates physical attention hijacking in the real world, although residual errors can still arise from remaining attention leakage to the patch under challenging sensing variations.

\FloatBarrier

\subsection{Ablation Study}
\label{subsec:ablation}

\begin{table}[!b]
  \centering
  \caption{SARF ablation on LIBERO-Spatial (Failure Rate, \%).
  Failure rates of OpenVLA under Clean, Random Patch, and AGSD Attack.
  Lower FR indicates better performance.}
  \label{tab:ablation_comp}
  \setlength{\tabcolsep}{6pt}
  \renewcommand{\arraystretch}{1.05}
  \scriptsize
  \begin{tabular}{l c c c}
    \toprule
    \multirow{2}{*}{\textbf{Method}} & \multicolumn{3}{c}{\textbf{Failure Rate (FR\%$\downarrow$)}} \\
    \cmidrule(lr){2-4}
    & \textbf{Clean} & \textbf{Random Patch} & \textbf{AGSD Attack} \\
    \midrule
    only $\mathcal{L}_{\text{pcad}}$ & 34.7 $\pm$ 2.0 & 41.1 $\pm$ 2.4 & 54.4 $\pm$ 2.5 \\
    only $\mathcal{L}_{\text{geo}}$  & 37.9 $\pm$ 1.5 & 59.8 $\pm$ 1.9 & 81.6 $\pm$ 1.2 \\
    \midrule
    w/o $\mathcal{L}_{\text{feat}}$ & 31.6 $\pm$ 2.5 & 34.9 $\pm$ 2.8 & 37.2 $\pm$ 3.0 \\
    w/o $\mathcal{L}_{\text{pcad}}$ & 14.2 $\pm$ 1.1 & 44.6 $\pm$ 2.2 & 87.5 $\pm$ 1.4 \\
    w/o $\mathcal{L}_{\text{geo}}$  & 14.0 $\pm$ 1.3 & 21.5 $\pm$ 1.6 & 35.8 $\pm$ 2.1 \\
    \midrule
    \textbf{Full SARF} & \textbf{14.4 $\pm$ 0.6} & \textbf{15.0 $\pm$ 0.7} & \textbf{17.0 $\pm$ 1.1} \\
    \bottomrule
  \end{tabular}
\end{table}

To validate the individual contribution of each SARF component towards defending against physical attacks and preserving original capabilities, we conduct an ablation study on LIBERO-Spatial (Table~\ref{tab:ablation_comp}). While Full SARF achieves the strongest robustness under Random Patch and AGSD Attack while maintaining near-original clean performance, removing $\mathcal{L}_{\text{pcad}}$ collapses robustness under AGSD (FR spikes to 87.5\%). Furthermore, dropping $\mathcal{L}_{\text{feat}}$ severely harms clean performance (31.6\% FR), and removing $\mathcal{L}_{\text{geo}}$ yields a moderate robustness drop (35.8\% FR). These results demonstrate that all three objectives are synergistic and indispensable: policy-critical attention correction drives the core defense, while feature anchoring and geometric consistency ensure the stability of the original visuomotor policy.

\FloatBarrier

\section{CONCLUSIONS}

We characterized physically realizable patch threats to Vision–Language–Action (VLA) manipulation and pinpointed a mechanism-level failure—policy-critical action-to-vision attention hijacking—that causally links localized visual perturbations to compounding control errors. Building on this diagnosis, we introduced AGSD as a stress-test physical attack that combines attention capture with vision–language semantic disruption, and we proposed SARF, a zero-inference-overhead defense that strengthens robustness by updating only the visual encoder while preserving the deployed policy interface. Across LIBERO suites and on a real PiPER platform, SARF consistently improves robustness under AGSD while maintaining clean-task performance, and attention visualizations corroborate that it re-centers action-conditioned attention onto task-relevant evidence during execution. The ablation results further show that policy-critical attention distillation is the dominant factor for attack robustness, while feature anchoring and geometric consistency help preserve clean-task stability. Overall, our findings suggest that mechanism-targeted robustness—stabilizing the attention pathways that directly drive action generation—offers a practical and deployable route to securing VLA robots against physical attacks without adding test-time complexity. In future work, we will extend SARF to natural clutter and distractors, further reduce residual long-horizon attention jitter, and validate the defense across broader VLA backbones, embodiments, and third-party adaptive attacks.






\bibliographystyle{IEEEtran}

\bibliography{refs}

@IEEEtranBSTCTL{BSTcontrol,
  CTLuse_forced_etal       = "yes",
  CTLmax_names_forced_etal = "3",
  CTLnames_show_etal       = "3"
}

@inproceedings{rt2,
  title={Rt-2: Vision-language-action models transfer web knowledge to robotic control},
  author={Zitkovich, Brianna and Yu, Tianhe and Xu, Sichun and Xu, Peng and Xiao, Ted and Xia, Fei and Wu, Jialin and Wohlhart, Paul and Welker, Stefan and Wahid, Ayzaan and others},
  booktitle={Conference on Robot Learning},
  pages={2165--2183},
  year={2023},
  organization={PMLR}
}

@article{openvla,
  title={Openvla: An open-source vision-language-action model},
  author={Kim, Moo Jin and Pertsch, Karl and Karamcheti, Siddharth and Xiao, Ted and Balakrishna, Ashwin and Nair, Suraj and Rafailov, Rafael and Foster, Ethan and Lam, Grace and Sanketi, Pannag and others},
  journal={arXiv preprint arXiv:2406.09246},
  year={2024}
}

@article{octo,
  title={Octo: An open-source generalist robot policy},
  author={Team, Octo Model and Ghosh, Dibya and Walke, Homer and Pertsch, Karl and Black, Kevin and Mees, Oier and Dasari, Sudeep and Hejna, Joey and Kreiman, Tobias and Xu, Charles and others},
  journal={arXiv preprint arXiv:2405.12213},
  year={2024}
}

@article{pi0,
  title={ $\pi_0$: A Vision-Language-Action Flow Model for General Robot Control},
  author={Black, Kevin and Brown, Noah and Driess, Danny and Esmail, Adnan and Equi, Michael and Finn, Chelsea and Fusai, Niccolo and Groom, Lachy and Hausman, Karol and Ichter, Brian and others},
  journal={arXiv preprint arXiv:2410.24164},
  year={2024}
}

@article{advpatch,
  title={Adversarial patch},
  author={Brown, Tom B and Man{\'e}, Dandelion and Roy, Aurko and Abadi, Mart{\'\i}n and Gilmer, Justin},
  journal={arXiv preprint arXiv:1712.09665},
  year={2017}
}

@inproceedings{eot,
  title={Synthesizing robust adversarial examples},
  author={Athalye, Anish and Engstrom, Logan and Ilyas, Andrew and Kwok, Kevin},
  booktitle={International conference on machine learning},
  pages={284--293},
  year={2018},
  organization={PMLR}
}

@article{mft,
  title={Manipulation facing threats: Evaluating physical vulnerabilities in end-to-end vision language action models},
  author={Cheng, Hao and Xiao, Erjia and Wang, Yichi and Yu, Chengyuan and Sun, Mengshu and Zhang, Qiang and Cao, Jiahang and Guo, Yijie and Liu, Ning and Xu, Kaidi and others},
  journal={arXiv preprint arXiv:2409.13174},
  year={2024}
}

@article{evavla,
  title={Eva-VLA: Evaluating Vision-Language-Action Models' Robustness Under Real-World Physical Variations},
  author={Liu, Hanqing and Long, Jiahuan and Wu, Junqi and Hou, Jiacheng and Tang, Huili and Jiang, Tingsong and Zhou, Weien and Yao, Wen},
  journal={arXiv preprint arXiv:2509.18953},
  year={2025}
}

@inproceedings{vla_vuln,
  title={Exploring the adversarial vulnerabilities of vision-language-action models in robotics},
  author={Wang, Taowen and Han, Cheng and Liang, James and Yang, Wenhao and Liu, Dongfang and Zhang, Luna Xinyu and Wang, Qifan and Luo, Jiebo and Tang, Ruixiang},
  booktitle={Proceedings of the IEEE/CVF International Conference on Computer Vision},
  pages={6948--6958},
  year={2025}
}

@article{edpa_vla,
  title={Model-agnostic adversarial attack and defense for vision-language-action models},
  author={Xu, Haochuan and Koh, Yun Sing and Huang, Shuhuai and Zhou, Zirun and Wang, Di and Sakuma, Jun and Zhang, Jingfeng},
  journal={arXiv preprint arXiv:2510.13237},
  year={2025}
}

@article{libero,
  title={Libero: Benchmarking knowledge transfer for lifelong robot learning},
  author={Liu, Bo and Zhu, Yifeng and Gao, Chongkai and Feng, Yihao and Liu, Qiang and Zhu, Yuke and Stone, Peter},
  journal={Advances in Neural Information Processing Systems},
  volume={36},
  pages={44776--44791},
  year={2023}
}

@article{diffpolicy,
  title={Diffusion policy: Visuomotor policy learning via action diffusion},
  author={Chi, Cheng and Xu, Zhenjia and Feng, Siyuan and Cousineau, Eric and Du, Yilun and Burchfiel, Benjamin and Tedrake, Russ and Song, Shuran},
  journal={The International Journal of Robotics Research},
  volume={44},
  number={10-11},
  pages={1684--1704},
  year={2025},
  publisher={Sage Publications Sage UK: London, England}
}

@article{upa_rfas,
  title={When Robots Obey the Patch: Universal Transferable Patch Attacks on Vision-Language-Action Models},
  author={Lu, Hui and Yu, Yi and Yang, Yiming and Yi, Chenyu and Zhang, Qixin and Shen, Bingquan and Kot, Alex C and Jiang, Xudong},
  journal={arXiv preprint arXiv:2511.21192},
  year={2025}
}

@article{advla,
  title={Attention-Guided Patch-Wise Sparse Adversarial Attacks on Vision-Language-Action Models},
  author={Zhang, Naifu and Tao, Wei and Xiao, Xi and Sun, Qianpu and Zheng, Yuxin and Mo, Wentao and Wang, Peiqiang and Zhang, Nan},
  journal={arXiv preprint arXiv:2511.21663},
  year={2025}
}

@article{attackvla,
  title={AttackVLA: Benchmarking Adversarial and Backdoor Attacks on Vision-Language-Action Models},
  author={Li, Jiayu and Zhao, Yunhan and Zheng, Xiang and Xu, Zonghuan and Li, Yige and Ma, Xingjun and Jiang, Yu-Gang},
  journal={arXiv preprint arXiv:2511.12149},
  year={2025}
}

@article{madry,
  title={Towards deep learning models resistant to adversarial attacks},
  author={Madry, Aleksander and Makelov, Aleksandar and Schmidt, Ludwig and Tsipras, Dimitris and Vladu, Adrian},
  journal={arXiv preprint arXiv:1706.06083},
  year={2017}
}

@inproceedings{trades,
  title={Theoretically principled trade-off between robustness and accuracy},
  author={Zhang, Hongyang and Yu, Yaodong and Jiao, Jiantao and Xing, Eric and El Ghaoui, Laurent and Jordan, Michael},
  booktitle={International conference on machine learning},
  pages={7472--7482},
  year={2019},
  organization={PMLR}
}

@inproceedings{domainrand,
  title={Domain randomization for transferring deep neural networks from simulation to the real world},
  author={Tobin, Josh and Fong, Rachel and Ray, Alex and Schneider, Jonas and Zaremba, Wojciech and Abbeel, Pieter},
  booktitle={2017 IEEE/RSJ international conference on intelligent robots and systems (IROS)},
  pages={23--30},
  year={2017},
  organization={IEEE}
}

@article{diffpure,
  title={Diffusion models for adversarial purification},
  author={Nie, Weili and Guo, Brandon and Huang, Yujia and Xiao, Chaowei and Vahdat, Arash and Anandkumar, Anima},
  journal={arXiv preprint arXiv:2205.07460},
  year={2022}
}

@article{defensegan,
  title={Defense-gan: Protecting classifiers against adversarial attacks using generative models},
  author={Samangouei, Pouya and Kabkab, Maya and Chellappa, Rama},
  journal={arXiv preprint arXiv:1805.06605},
  year={2018}
}

@article{pixeldefend,
  title={Pixeldefend: Leveraging generative models to understand and defend against adversarial examples},
  author={Song, Yang and Kim, Taesup and Nowozin, Sebastian and Ermon, Stefano and Kushman, Nate},
  journal={arXiv preprint arXiv:1710.10766},
  year={2017}
}

@inproceedings{patchguard,
  title={$\{$PatchGuard$\}$: A provably robust defense against adversarial patches via small receptive fields and masking},
  author={Xiang, Chong and Bhagoji, Arjun Nitin and Sehwag, Vikash and Mittal, Prateek},
  booktitle={30th USENIX Security Symposium (USENIX Security 21)},
  pages={2237--2254},
  year={2021}
}

@inproceedings{obfg,
  title={Obfuscated gradients give a false sense of security: Circumventing defenses to adversarial examples},
  author={Athalye, Anish and Carlini, Nicholas and Wagner, David},
  booktitle={International conference on machine learning},
  pages={274--283},
  year={2018},
  organization={PMLR}
}

@article{ewc,
  title={Overcoming catastrophic forgetting in neural networks},
  author={Kirkpatrick, James and Pascanu, Razvan and Rabinowitz, Neil and Veness, Joel and Desjardins, Guillaume and Rusu, Andrei A and Milan, Kieran and Quan, John and Ramalho, Tiago and Grabska-Barwinska, Agnieszka and others},
  journal={Proceedings of the national academy of sciences},
  volume={114},
  number={13},
  pages={3521--3526},
  year={2017},
  publisher={National Academy of Sciences}
}

@article{lwf,
  title={Learning without forgetting},
  author={Li, Zhizhong and Hoiem, Derek},
  journal={IEEE transactions on pattern analysis and machine intelligence},
  volume={40},
  number={12},
  pages={2935--2947},
  year={2018},
  publisher={IEEE}
}

@article{attntransfer,
  title={Paying more attention to attention: Improving the performance of convolutional neural networks via attention transfer},
  author={Zagoruyko, Sergey and Komodakis, Nikos},
  journal={arXiv preprint arXiv:1612.03928},
  year={2016}
}

@article{retain_merge,
  title={Robust Finetuning of Vision-Language-Action Robot Policies via Parameter Merging},
  author={Yadav, Yajat and Zhou, Zhiyuan and Wagenmaker, Andrew and Pertsch, Karl and Levine, Sergey},
  journal={arXiv preprint arXiv:2512.08333},
  year={2025}
}

@inproceedings{cai2025mft_fusion,
  title={MFT: Modal Fusion Transformer for Cross-Modal Fusion in 3D Object Detection},
  author={Cai, Haojie and Yin, Dongfu and Yu, Fei and Xiong, Siting},
  booktitle={ICASSP 2025--2025 IEEE International Conference on Acoustics, Speech and Signal Processing (ICASSP)},
  pages={1--5},
  year={2025},
  organization={IEEE}
}

@inproceedings{cai2025dstr,
  title={DSTR: Dual Scenes Transformer for Cross-Modal Fusion in 3D Object Detection},
  author={Cai, Haojie and Yin, Dongfu and Yu, Fei Richard and Xiong, SiTing},
  booktitle={2025 IEEE/CVF Winter Conference on Applications of Computer Vision (WACV)},
  pages={3064--3073},
  year={2025},
  organization={IEEE}
}

\end{document}